\documentclass[sigconf]{acmart}
\usepackage{multirow}
\usepackage{algorithm}
\usepackage{amsmath}
\usepackage{amssymb}
\usepackage[noend]{algpseudocode}

\def\BibTeX{{\rm B\kern-.05em{\sc i\kern-.025em b}\kern-.08emT\kern-.1667em\lower.7ex\hbox{E}\kern-.125emX}}

\renewcommand\footnotetextcopyrightpermission[1]{} 
\setcopyright{none}

\begin{document}

\title{Coverage Testing of Deep Learning Models using Dataset Characterization}

\author{Senthil Mani}
\affiliation{IBM Research}

\author{Anush Sankaran}
\affiliation{IBM Research}

\author{Srikanth G Tamilselvam}
\affiliation{IBM Research}

 \author{Akshay Sethi}
\affiliation{Borealis AI}
\authornote{Akshay Sethi was a part of IBM Research when this work was performed.}

\renewcommand{\shortauthors}{Mani, et al.}

\begin{abstract}
Deep Neural Networks (DNNs), with its promising performance, are being increasingly used in safety critical applications such as autonomous driving, cancer detection, and secure authentication. With growing importance in deep learning, there is a requirement for a more standardized framework to evaluate and test deep learning models. The primary challenge involved in automated generation of extensive test cases are: (i) neural networks are difficult to interpret and debug and (ii) availability of human annotators to generate specialized test points.

In this research, we explain the necessity to measure the quality of a dataset and propose a test case generation system guided by the dataset properties. From a testing perspective, four different dataset quality dimensions are proposed: (i) equivalence partitioning, (ii) centroid positioning, (iii) boundary conditioning, and (iv) pair-wise boundary conditioning. The proposed system is evaluated on well known image classification datasets such as MNIST, Fashion-MNIST, CIFAR10, CIFAR100, and SVHN against popular deep learning models such as LeNet, ResNet-20, VGG-19. Further, we conduct various experiments to demonstrate the effectiveness of systematic test case generation system for evaluating deep learning models.
\end{abstract}



\maketitle

\section{Introduction}
Over the past few years, Deep Neural Networks (DNNs) have made significant progress in many cognitive tasks such as image recognition, speech recognition, and natural language processing. Availability of large amounts of unlabeled training data has enabled deep learning from achieving near human accuracy in many day-to-day tasks. This promising technology development has empowered deep learning to be increasingly used in safety critical production-ready applications such as self-driving cars \cite{angelova2015real}~\cite{bojarski2016end}~\cite{huval2015empirical}, flight control systems \cite{zhang2016learning}, and medical diagnosis~\cite{esteva2017dermatologist}~\cite{milletari2016v}.

Currently, the accuracy of a deep learning model computed on the test set is the common metric used for measuring the overall performance of the model. However, this could be insufficient because:

\begin{enumerate}
    \item In most of the popular datasets, the stand out test dataset is typically handpicked or randomly chosen from the entire dataset 
    \item The provided test data may not be a true representative of the data obtained in real world
    \item The test data set may not have a good coverage of the data distribution the model is trained on
\end{enumerate}

The quality of the test data set is an important factor which influences the acceptance of the accuracy metric of the model evaluated. If the test data set in true sense does not represent the production data, or real world data, or is biased, then accuracy of the model reported cannot be trusted. 

\begin{figure*}[t]
	\begin{center}
	\includegraphics[width=.95\textwidth]{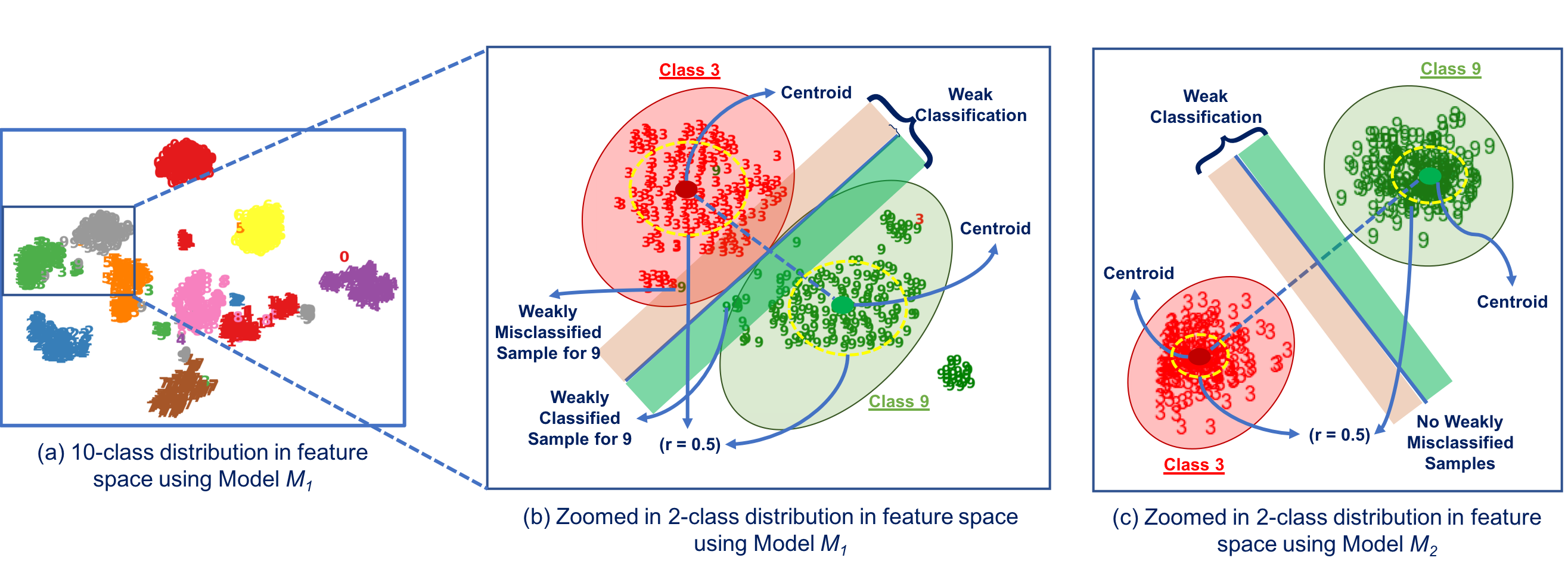}
	\end{center}
	\caption{Feature space representation of data points (images) classified as either class 3 or 9.}
	\label{fig:approach}
\end{figure*}

Traditional programs are deterministic and hence exhaustive coverage analysis was a tractable solution. However, DNNs are data driven and hence, standard approaches for testing the model is to gather real world test data as much as possible. Such datasets are manually labelled in a crowd sourced manner~\footnote{\url{https://www.figure-eight.com/}}~\footnote{\url{https://www.mturk.com/}} which is a costly and time consuming process~\cite{fei2010imagenet} \cite{GoogleFinanceMonitise20140426}. Also, different DNNs based on their complexity perceive the data differently i.e their classification boundaries tend to be different. Therefore, there is a need to explore the input data space, and test data generation based on the architecture details and the complexity of the  model. Otherwise, the coverage of the model would be incomplete and the model may not be a true representative of real world application.

Consider a simple classification algorithm, which is trained on a MNIST multi-class dataset. It is basically a set of numbers as images, which needs to be classified as 0 through 9 (10 classes). The test dataset ideally, should have test cases sampled across all these classes with no distribution bias (equally sampled across labels 1 though 9). Further, it should contain test cases where the images are very clear representations of the numbers and also images, where the numbers look like they are overlapping with other classes. For example, images containing number 1 and number 7 can overlap significantly because of how the numbers are written, as there is high overlap among the strokes. Similarly, there will be very little overlap among numbers like 1 and 8. Hence, the test dataset should contain appropriate samples (images) from both overlapping and non-overlapping classes. If such a test dataset can be constructed which can be considered to have a good coverage across the entire distribution of the data set, then the accuracy number reported on this test set can be trusted.

Figure \ref{fig:approach} provides a representation of the feature space of the data sets classified as labels 3 and 9 by two different DNN models $M_1$ and $M_2$. A feature space is a collection of features related to some properties of the object and the number of features determines the dimensionality of the space. There are always data points (images) in the observed feature space, which are clearly classified as either of the classes. They tend to be closer to the centroid of their respective cluster. However, there will also be data points (images) which are in the boundary, which could either be weakly classified or miss-classified. In this case, the test dataset when observed in the feature space of $M_1$, the data points are spread across the space, some closer to the centroids and some near the boundaries. However model $M_2$, has clearly classified all the points in the same test dataset to either classes, closer to the centroid.

We hypothesise that the accuracy obtained using this test dataset using model $M_1$ is \textit{more guaranteed} or \textit{trustworthy}, since the test dataset had a broader coverage of data points in the feature space, than model $M_2$. 
Given the importance of test dataset in validating the model, in this research, we propose the following four metrics to measure the goodness of test dataset based on the coverage of the data points in the feature space of the model, and further use these dimensions to guide generation of ideal test datatset, on which the model's evaluation is more guaranteed or trustworthy.  
\begin{enumerate}
    \item \textbf{Equivalence partitioning}: Measures the distribution of test data across individual classes.
    \item \textbf{Centroid positioning}: For each class, measures the percentage of test data that lie close to the centroid of the trained class cluster
    \item \textbf{Boundary conditioning}: For each class, measures the percentage of test data that lie near the boundary with respect to every other class of trained class clusters 
    \item \textbf{Pair-wise boundary conditioning}: Measure for each pair of class the percentage of the boundary conditioning
\end{enumerate}

To the best of our knowledge, there is no existing work which proposed coverage of the data points in the feature space for testing the model. 

\noindent We use these metrics to measure the quality of the test dataset in the feature space and use it for sampling additional test cases. We empirically evaluate and present the results of goodness of the original test samples of five popular image datasets: \textit{MNIST}~\cite{lecun1998gradient}, \textit{FashionMNIST (FMNIST)}~\cite{xiao2017fashion}, \textit{CIFAR-10}~\cite{krizhevsky2009learning}, \textit{CIFAR-100}~\cite{krizhevsky2009learning}, and \textit{SVHN}~\cite{netzer2011reading} and on three popular state of the art deep learning models: \textit{VGG-19}~\cite{simonyan2014very}, \textit{LeNet}~\cite{lecun2015lenet}, and \textit{ResNet-20}~\cite{he2016deep}. Models which which were evaluated on benchmark data sets, which were also ideal test dataset based on our metrics, showed minimum variance in accuracy when tested with test datasets sampled across the dimensions. However models such as \textit{VGG-19} and \textit{ResNet-20}, which had reported high accuracy ranges on the benchmark test dataset which were not ideal as measured by our metrics, showed a significant drop in accuracy (of more than $70\%$) when tested on the ideal test dataset.

To summarize, the main contributions of our paper are :
\begin{itemize}
  \item  A set of four metrics to measure the coverage quality of test dataset, in the feature space of the model. 
  \item  A guided systematic approach to sample additional test cases from the feature space. 
  \item  An empirical study on the quality of most common test datasets on popular deep neural network models 
\end{itemize}

The rest of the paper is organized as follows. The prior literature is discussed in Section 2. Section 3 provides a background on deep neural networks. The details of our metrics and approach are explained in Section 4. Section 5 details the experiment setting and the evaluation. Section 6 provides a discussion and limitation of our approach followed by conclusion and future work in Section 7.


  
  
  

\section{Existing Literature}
We categorize the existing set of related research works in the literature into three categories: (i) testing based on model and data coverage, (ii) adversarial testing methods, and (iii) metrics based testing. 

\subsection{Coverage Based Testing} 
Pei et al., \cite{pei2017deepxplore}, first introduced neuron coverage as a metric for testing DNN models. Neuron coverage of a DNN can be compared to code coverage of traditional systems which measures the extent of code exercised by the input sample. Test dataset that result in every hidden unit getting activated i.e., positive value for atleast one of the input test sample are considered to have complete coverage. Then multiple DNNs are cross referenced using gradient based optimization to identify erroneous boundary cases. In a way, it is claimed that test dataset that gets full or high neurons activation can be considered good quality. 
    
Ma et al., \cite{ma2018deepgauge}, estimate the testing adequacy of DNNs in two ways. In Major Function behavior, the activation values are checked if they fall within the minimum and maximum neuron activation values observed during training. Further, an in-depth coverage analysis called k-multi section neuron coverage is done by partitioning the region into k sections between the boundaries, and measure if each of them have been visited. In Corner Case behavior, they measure whether each activation goes beyond or below a certain boundary. It is claimed that test datasets whose neuron activation values spread across the k boundaries and close to the corner regions can be considered good quality.
    
Sun et al. \cite{sun2018testing}, introduced four different test criteria inspired from Modified Condition/ Decision Coverage \cite{hayhurst2001practical}. Adequacy, as a measure, is also covered here, however, interestingly their criteria also studies the effects of features from the adjacent layer. Their intent comes from the fact that deeper neural layer capture complex features and therefore its next layer can be considered as its summary.

Tian et al. \cite{tian2018deeptest} generated synthetic test images by applying traditional transformations such as blurring, shearing etc to maximize neuron coverage. They then tested erroneous behavior using metamorphic relations.
    
Odena et al. \cite{odena2018tensorfuzz} 
measure the model coverage by looking at the activations of computation graph. They proposed a coverage guided mutation techniques to mutate the inputs towards the goal of satisfying user-specified constraints.
    
Additionally, Sun et al. \cite{sun2018testing} claim that neuron coverage is a coarse criterion and it is easy to find a test dataset that achieves 100\% coverage. To demonstrate, they randomly picked 25 images from MNIST test set and for each of the test sample, if a neuron is not activated, sampling its value from [0, 0.1] gave them complete coverage.  Therefore, we focus on higher dimension space to study coverage and boundary conditions of test set and use them for guided test dataset generation. 

\subsection{Adversarial Testing} 
Ma et al. \cite{ma2018deepmutation} proposed few operators to introduce changes both at data and model level and evaluated
quality of test data by analyzing the
extent to which the introduced changes could be detected.
Similarly many existing works \cite{szegedy2013intriguing} \cite{nguyen2015deep} \cite{moosavi2017universal} \cite{carlini2017towards},
apply various heuristics, mostly based on gradient
descent or evolutionary techniques modify the important pixels. These approaches may
be able to find adversarial samples efficiently, however, does not guarantee about the existence of non-adversarial test examples.

\subsection{Metrics Based Testing} 
Most of the current DNN models rely just on the prediction accuracy (similar to black-box system testing that compares inputs and its corresponding outputs), lacking systematic testing coverage criteria. This is not sufficient which is further shown by the surge in different testing methods such as Concolic testing \cite{sun2018concolic}.

\subsection{Challenges in Deep Neural Network Testing}
As shown in this section, there are multiple research works studying the importance of testing DNN models. However, there are a couple of broad level challenges and research gaps in the existing literature, as summarized below:

\begin{enumerate}
    \item \textbf{Model Specific Testing:} Test dataset evaluation and test dataset generation has to be customized for the models being tested. There is limited amount of research in model specific test dataset quality estimation and generation
    \item \textbf{Feature Space Engineering:} Most of the testing techniques aim at transforming in the input data space (directly altering images or text). There is little work in understanding the latent features learnt by the model and generating additional test cases based on that.
    \item \textbf{Model Verification:} The primary challenge with test case generation is to define the ground truth oracle for each of the generated test sample. However, there is little efforts in creating a rule book of test case to verify the properties of the model.
\end{enumerate}

\begin{figure*}[t]
	\begin{center}
		\includegraphics[width=0.95\textwidth]{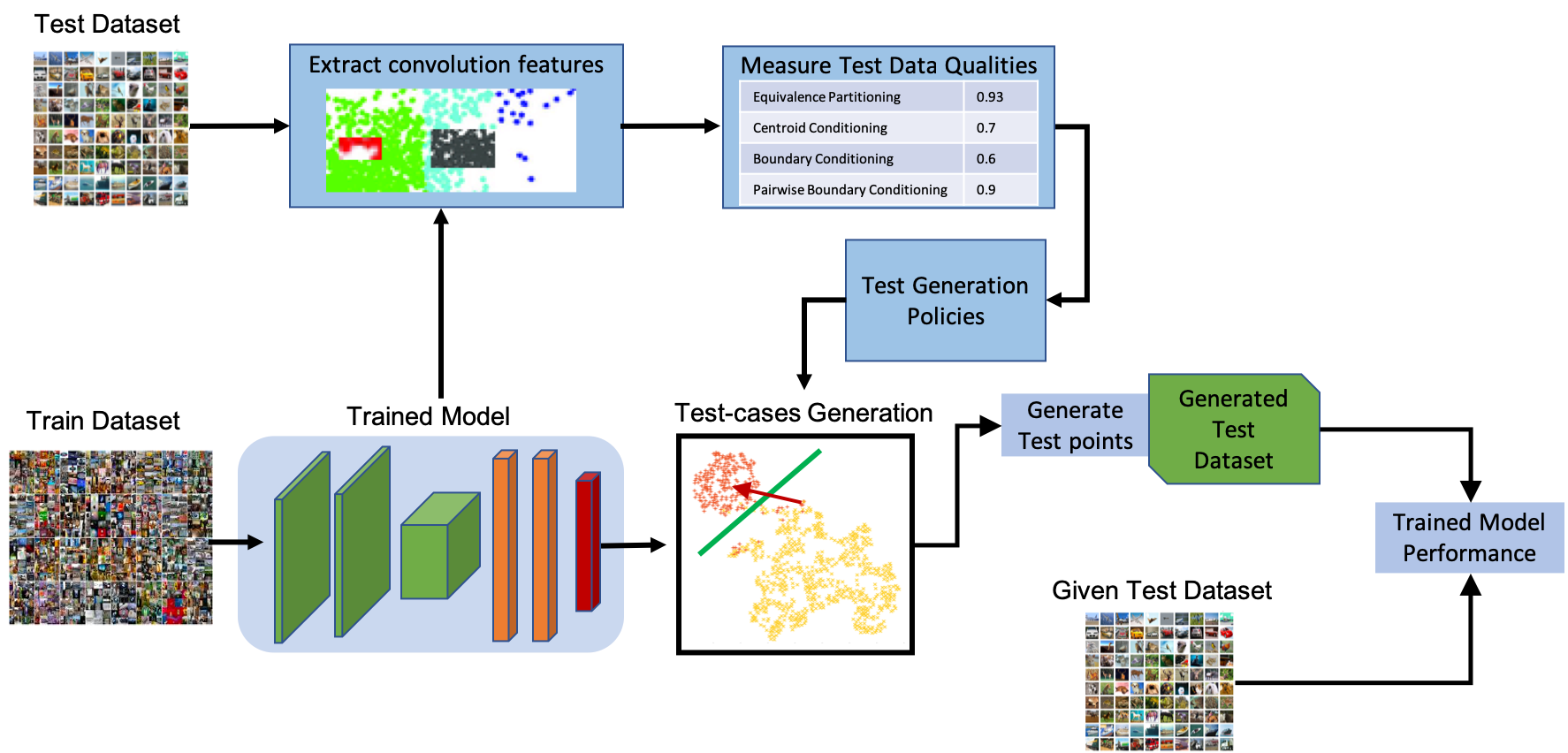}
	\end{center}
	\caption{The outline of the proposed approach explaining the overall framework for test dataset quality measurement based test dataset generation. A neural network model is trained using the train dataset. Then on the test dataset, image features are extracted from the trained model's last layer. These four measures are studied on these features based on which test cases are generated. The model's performance is now tested for this guided test dataset.}
	\label{fig:framework}
\end{figure*}

\section{Deep Neural Network: Overview}
Inspired from the functioning of a human brain, a deep neural network consists of a sequence of layers which converts the input signal into a task. Deep neural networks, with its sequence of nonlinear transformations, are known to learn highly robust and discriminative features for the given data and task~\cite{sun2018testing}. There are different kinds of DNN architectures~\cite{tian2018deeptest}: (i) Fully connected feed forward neural network, (ii) Convolutional neural network (CNN), and (iii) Recurrent Neural Network. A feed forward neural network works with numerical or categorical data as input. A CNN is a special type of neural network which takes multi-dimensional image data as input while RNN works on sequential time-series input data. 

The primary difference between a feed forward neural network and a CNN is the presence of a convolutional layer. Each convolutional layer has a small group of learnable neurons (filters), where each filter extracts some features from the image. The primary advantage of the convolutional layer is sharing weights, ie, the neurons in the convolutional layer are connected to only a few neurons in the previous layer, thereby drastically reducing the number of weight parameters 
A CNN network consists with a sequence of operations such as convolutional layer, pooling layer, fully connected layers, and activation layer. Consider an image, \textit{I}, the convolutional operation is shown as follows,
\begin{equation}
Conv(I) = \coprod_{i=1}^{n}  (I \circledast w_i)
\end{equation}
where, $w_i$ is a small size square filter, typically of size $3\times3$, $5\times5$, or $7\times7$, $\circledast$ is the convolutional operation, and $\coprod$ is the concatenation of the $n$ different filter responses. A typical CNN model consists of a sequence of operations (typically, $20$ - $150$) such as,
\begin{multline}
    CNN(I) = Softmax( Dense_2 (Dense_1( \\ 
    ReLU_3 (Pool_3 (Conv_3 ( \\ ReLU_2 (Pool_2 (Conv_2 ( \\ReLU_1 (Pool_1 (Conv_1 (I))) ))) )))
\end{multline}
where, $Pool$ is a Pooling2D operation, $ReLU$ and $Softmax$ are non-linear activation operations, and $Dense$ is a fully connected layer. Each of these operations (also called, layers) can be viewed as extracting different features from input image. It is a well established concept that the initial few layers learns coarse high level features and the terminal few layers learns low level features~\cite{shin2016deep}.

\section{Quality Estimation of Test Data}

A Convolutional Neural Network (CNN) could be considered as any other software system, with the program flow in a software system equivalent to the data flow in the CNN. CNNs are typically used for classification task like object classification, object detection among others. Classification task can be either binary class (simple yes or no) or multi-class (like numbers 0 through 9). When the CNNs are trained on the dataset, they learn some features automatically to fit a non-linear boundary to classify the group of data sets, based on the ground truth label. Depending on the complexity of the models, the number and type of layers,  hyper-parameters, and number of iterations the model was trained on, each model will learn a different non-linear boundary to classify on the same data set. Hence, a standard test data set when inferred through these different models, might place the same data point anywhere in the feature space: on the boundary, close to the boundary or farther away from it. 

Coverage techniques like neuron coverage \cite{tian2018deeptest}, can help test the model focusing on coverage of the number of neurons in each layer of the model (similar to  statement coverage in a traditional software program). However from a data coverage perspective a standard test data set does not suffice. Depending on the model, and the learnt representation of the data in the feature space, we need to sample the data points to have a coverage on the input data space. To the best of our knowledge, there is no existing work which proposes coverage of the data points in the feature space for testing the model. 

Further the measure of accuracy as a performance evaluation of the entire model on a standard data set, is not trust worthy. The accuracy only holds, if data in the wild is similar to the distribution of the test data set on which it was evaluated \cite{torralba2011unbiased}. Hence, if the test data set did not have enough coverage on the input data space, then the accuracy reported is very narrow and is not a general or a broader representation of the model performance. 



In this paper, we propose an approach which leverages the learned representations and the classification boundaries for evaluating the quality of the test set. The fundamental intuition is that the test set should be well spread in the feature space so as to do a maximum coverage of systematically testing the model's performance. 

\subsection{Properties of a Classifier}
As shown in Figure~\ref{fig:approach}, a well trained deep learning model has the following properties.

\begin{itemize}
  \item \textit{Centroid} of a class is the mean representation of the spread of the class data points in the feature space. Hence as we sample data points along the line from one class \textit{centroid} towards another class \textit{centroid} there should be a decrease in the class probability of former class and increase in the class label probability of the latter, predicted by the model.  
  \item Exploiting the boundary conditions between the two classes, we should be able to identify weakly misclassified points. When moving from these weak misclassifications towards the centroid of ground truth class, the probability of incorrectly predicted class by the model should decrease and probability of the correct class should increase.
  \item Finally for each data point in the test dataset, the probability of the ground truth class should ideally be more than any other class.
\end{itemize}
It is to be noted that a CNN model has a sequence of multiple complex non-linear transformation of the input image. The output of each layer constitutes an independent feature space that are non linearly correlated with the feature spaces obtained from the other layers. However, the above properties are applicable for all the intermediate feature spaces of the deep learning model. 

\subsection{Metrics for Evaluating Test Data Set Quality}
Based on the properties discussed, we propose four metrics for measuring quality of a test data set.

\textbf{(1) Equivalence Partitioning}\label{equi_partition}
This measures the distribution of test samples across all the classes. The hypothesis is that the test data set should contain equally distributed test samples from all the classes to avoid any bias in the testing the model towards any subset of classes. We measure the class level equivalence in the test data set as follow,
     \begin{equation}
         \text{Equivalence partitioning}, EP_i  = \frac{(ns_i * nc)}{ns}   
     \end{equation}
where, $ns_i$ is the number of test samples belonging to class $i$, $nc$ is the total number of classes, and $ns$ is the total number of samples in the test set. The ideal score is expected to be close to 1 for all classes.
       
\textbf{(2) Centroid Positioning}
This measures the number of test samples that lie in the centroid region of the class cluster spread. The hypothesis is the test cases should be equally well spread in the feature space of the model. The centroid region of a class is calculated by averaging out all the features vectors of points belonging to a single class. The normalized euclidean distance of all the points belonging to the class are obtained and a radius threshold of $r$ is used to classify whether the test point is in the centroid region. The specific threshold value used to measure is explained in our experiment section. The centroid positioning score of a particular class of test data is computed as follows.

\begin{multline}
	\text{Centroid Positioning}, CP_i  = \frac{\sum_{j=1}^{ns_i} cent(ns_i^{(j)})}{ns_i}     \\
	\text{where,}
	cent(x) = \begin{cases}
		1,  & \text{if } dist(x, centroid) \leq r\\
		0,  & \text{otherwise}
	\end{cases}
\end{multline}

The obtained score is bounded in the range of [0,1] where the ideal score should tend towards 0 for each class.

\textbf{(3) Boundary Conditioning} 
The aim here is to measure the number of test data points that are towards the classification boundary. The region near the boundaries are those with maximum confusion for the classifiers and hence testing in this region would provide a robust evaluation of the model. In an ideal scenario, there is a need for maximum number of test points with a good distribution to lie near the boundary. Thus, test samples with confidence in the range of $[\theta\textunderscore1, \theta\textunderscore2]$ are considered as weakly classified samples that lie near the boundary. $[\theta\textunderscore1, \theta\textunderscore2]$ values are explained in our experiment section.

\begin{multline}
     \text{Boundary Conditioning}, BC_i  = \frac{\sum_{j=1}^{ns_i} bound(ns_i^{(j)})}{ns_i}     \\
     \text{where,}
     bound(x) =   \begin{cases}
        1,  & \text{if } confidence(x) \in  [\theta\textunderscore1, \theta\textunderscore2]\\
        0,  & \text{otherwise}
    \end{cases}
\end{multline}


\textbf{(4) Pairwise Boundary Conditioning} 
This measures the boundary conditioning for every pair of classes. This measure is used to check if the boundary conditions are equally tested for all pair of classes in the dataset.

\subsubsection{Existing Metrics}
There are studies in the literature that discusses different metrics to measure the quality of the test dataset and also the impact of the test dataset quality on the performance of a machine learning model. Turhan~\cite{turhan2012dataset} studied the goodness of a test dataset as a dataset shift problem. The basic hypothesis is that the distribution of the test dataset should neither be too far away nor too overlapping with the train dataset. A highly divergent test dataset would test a machine learning prediction model on a feature space that it was not trained on, resulting in poor testing and results. Also, a highly overlapping test dataset would not test the model on its generalization capability. 

Specifically, a simple covariate shift~\cite{storkey2009training} has been used as a popular metric to study the impact on test data on machine learning prediction models. For a given dataset $\textbf{x}$ with labels $y$, a machine learning model $P(y|\textbf{x})P(\textbf{x})$ is learnt. Covariate shift occurs when the covariates of the test data, $P(\textbf{x}_{test})$ differs from the train data, $P(\textbf{x}_{train})$. A common example in image datasets could be that the different images of an object (say, airplane) are captured during the day time in the train dataset. While in the test dataset, the same objects are captured during the night time (with dark background) shifting the properties of the test dataset away from the train dataset. 

However, these metrics does not take into consideration the coverage criteria to measure the quality of the test dataset. Additionally, these metrics are model agnostics and does not include the characteristics and the complexity of the model. In this research, we postulate that the test datasets which has a good quality measure using these existing metrics can still suffer in terms of coverage and model dependent testing. Thus, we require additional metrics to measure the quality of test datasets.

\begin{algorithm}[!t]
\caption{test dataset Generation}\label{algo:generation}
\begin{algorithmic}[1]
\State $r \gets$ Set centroid positioning threshold
\State $wc_l \gets$ Set weak class lower boundary confidence value less than $\theta_2$
\State $d \gets$ list [0, 20, 30, 50, 70, 80, 100] \ of\  test dataset $distribution$ choices in \% 
\State $f_c \gets$ list [10, 25, 50, 75, 100] \ of\ test dataset $frequency$  choices in \%
\For{\texttt{Dataset $K$ in Datasets}}
\State $Model \gets$ Load DNN model trained on $K$
\State $c_i \gets$ Number of samples of class $i$
\State $c_c \gets$ Calculate centroid of the class $i$
\State $cp_i  \gets$ Measure centroid positioning i.e i samples that fall inside $r$ of $c_c$ 
\State $bc_i  \gets$ Measure boundary condition for i samples i.e  $ <= wc_l$

\State $f_k \gets$ pick k from $f_c$
\For{\texttt{ For $d_c$ in $d$}} 
          \State $d_i \gets$ \Call{GENERATE}{$i$, $d_c$, $f_k$}
          \State Return $d_i$ 
\EndFor
\EndFor 

\\
\Procedure{GENERATE}{$i$, $d_c$, $f_k$}
\State Select all $bc_i$ samples of $i$ 
\State Perturb $bc_i$ to generate samples $s_i$ optimizing $c_i$ , $bc_i$ w.r.t to $x$ distribution and $f_k$ count constraint
\State Apply DeConv to obtain images from features $s_i$
\State Return $d_i$
\EndProcedure
\end{algorithmic}
\end{algorithm}

\subsection{Test Data Generation}
The overall approach used to generate test data set is illustrated in Figure~\ref{fig:framework}. Given a test dataset and model trained on it, the class-wise quality score using the metrics are evaluated. These scores provides an insight on what region of the data set has not been well represented in the test data set for a given model. We use these insights as guidance for generation of test samples in the feature space. Further, these sampled features are given to a trained deconvolutional network to visualize the actual image in the data space. Deconvolutional or Transpose Convolution Network is a common technique for learning upsampling of an image. This network takes a feature representation and reproduces the original image. Our deconvolution follows the same architecture as \cite{dosovitskiy2016inverting}. We use such a network to generate samples which are human recognizable images representation of the features.

However, we used the boundary conditioning property, to calculate the accuracy metric for test samples in the feature space close to boundary for which a meaningful image is not generated by deconvolutional network. 

Algorithm \ref{algo:generation} explains the step-by-step procedure for test data set generation. For a given test dataset $K$, the quality measurement is extracted using all the four proposed metrics. In the next step, we generate additional test samples driven by the extracted quality measurements, with the motive of expanding the test dataset coverage. To generate additional samples in the features space, we experimented with different distribution choices, $d$ and different frequency choices $f_c$. Depending on the centroid positioning value $cp_i$ and the boundary condition value $bc_i$, the existing points in the features are perturbed to generate new test samples. The perturbation is performed in a controlled manner to ensure that the new test samples remain in the boundary or towards the centroid. The generated test dataset which complements the coverage of the original test dataset is then returned to measure the model\'s guaranteed performance.

         


The primary research questions that we study and experimentally analyze in this research paper are as follows:
\begin{enumerate}
    \item \textbf{RQ1:} Does the existing data set specific metrics sufficiently describe the quality of the test set?  
    \item \textbf{RQ2:} Does our proposed metrics sufficiently describe the quality of the test set?
    \item \textbf{RQ3:} Does the generated additional test dataset provide a more ``guaranteed" measure of the model's performance?
\end{enumerate}


\section{Experimental Results and Analysis}

In this section, we provide details of the different publicly available datasets and existing models that are used for the experiments. The results are provided for all the three proposed research questions and analyzed. We also discuss the implications of our results to the research community.

\subsection{Datasets and Models}
To experimentally evaluate our approach of test data quality determination and guided test case generation we use five standard vision benchmark datasets: (i) \textit{MNIST}, (ii) \textit{F-MNIST}, (iii) \textit{CIFAR-10}, (iv) \textit{CIFAR-100}, and (v) \textit{SVHN}. One each of these datasets we run three diverse and popular CNNs to study the feature spaces created by multiple models: (i) \textit{LeNet}, (ii) \textit{VGG-19}, and (iii) \textit{ResNet-20}. LeNet is a basic and one of the first CNN to be proposed with 5 trainable layers. \textit{VGG-19} is a \textit{de facto} baseline CNN model with $19$ trainable layers. \textit{ResNet-20} is a 20 layer network and one of the popular state-of-art the models in different image classification applications. Table~\ref{table:dataprop} shows the properties of the five different datasets and the accuracy of these three models on each of the dataset, computed using the benchmark train and test sets. 
On all of these dataset and model combination, we study the quality of the standard test set that is provided as a part of the respective benchmark dataset.

\begin{table}[!t]
	\centering
	\begin{tabular}{|l|l|l|l|l|l|l|}
		\hline
		\multirow{2}{*}{Dataset}  &
		\multirow{2}{*}{Class} & \multirow{2}{*}{\#Train} & \multirow{2}{*}{\#Test} & \multicolumn{3}{|c|}{Accuracy (\%)} \\ \cline{5-7}
		  &  &  &  & LeNet & VGG & ResNet \\ 
		\hline
		MNIST & 10 & 60000 & 10000 & 99.49 & 99.61 & 99.60   \\ \hline
		FMNIST & 10 & 60000 & 10000 & 88.50 & 93.13 & 92.58    \\ \hline
		CIFAR10 & 10 & 50000 & 10000 & 70.67 & 91.00 & 92.43    \\ \hline
		CIFAR100 & 100 & 50000 & 10000 & 37.23 & 61.38 & 67.41   \\ \hline
		SVHN & 10 & 73257 & 26032 & 89.50 & 96.80 &  96.40  \\ \hline
	\end{tabular}
	\caption{Properties of the five different image datasets used in our experiments.}
	\label{table:dataprop}
\end{table}

\begin{figure}[!t]
	\begin{center}
	\includegraphics[width=0.48\textwidth]{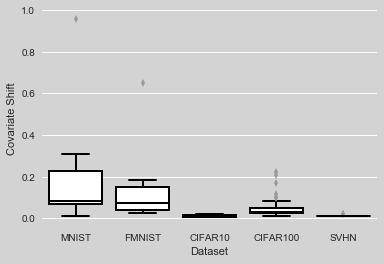}
	\end{center}
	\caption{Box plot showing the covariance shift~\cite{turhan2012dataset} of the test dataset with respect to the train dataset the across the classes for each dataset. The covariance shift is normalized between [0,1] where the $0$ represents that the test data is sampled exactly from the distribution of the train data inferring good quality.}
	\label{fig:cov_shift}
\end{figure}

\begin{figure}[!t]
	\begin{center}
	\includegraphics[width=0.48\textwidth]{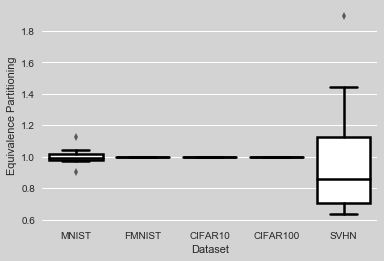}
	\end{center}
	\caption{The value of equivalence partitioning (EQ) for each class across all the five datasets.}
	\label{fig:eq_part}
\end{figure}

\begin{figure*}[!t]
	\begin{center}
	\includegraphics[width=0.9\textwidth]{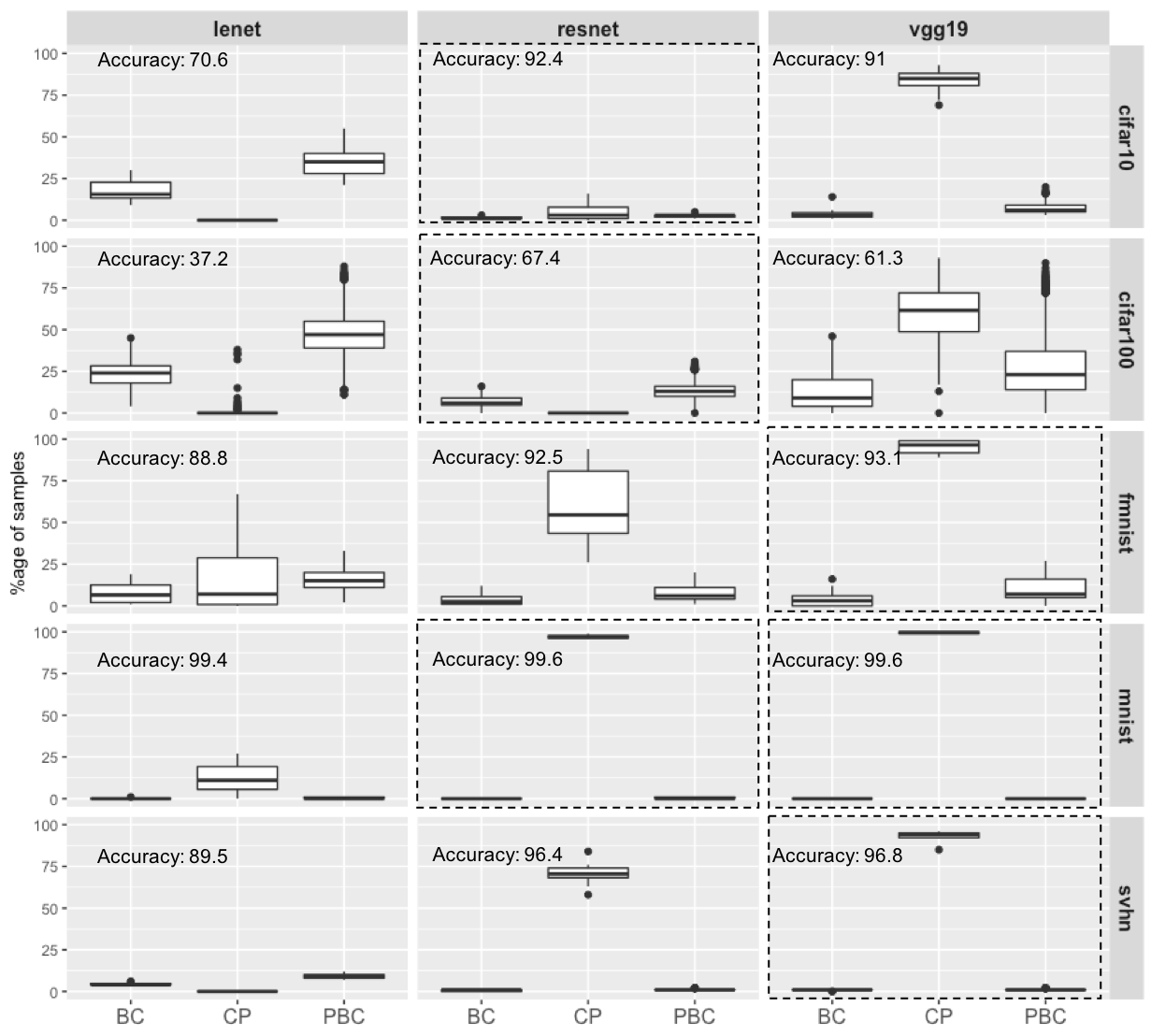}
	\end{center}
	\caption{Lists the box-plots of three models (LeNet, ResNet and VGGNet) evaluated against the test datasets of 5 data sets (cifar-10, cifar-100, fmnist, mnist and svhn). Each plot shows the distribution of test samples (percentage) across three dimensions names - border (BC), centroid (CP) and pair-wise border (PBC). Models which have the highest accuracy for the data set has been highlighted with dashed-border. The models' accuracy is also mentioned in each of the plots. }
	\label{fig:expts}
\end{figure*}

\subsection{RQ1: Quality Analysis using Existing Metrics}
In this experiment, we study the quality of the existing benchmark test sets using the existing quality metrics discussed in section \textit{4.2.1}. 

As explained by Turhan~\cite{turhan2012dataset}, the covariane shift of the test datase with respect to the train dataset is measured. For every class in every dataset, a Gaussian Mixture Model (GMM) is fit with number of components $10$. The dataset shift is then measured using Jensen-Shannon divergence between the $GMM_{train}$ model and $GMM_{test}$ model. For each dataset and each class in the dataset, the divergence meeasure is computed and it shown in Figure~\ref{fig:cov_shift}. 

It can be observed that for \textit{MNIST} dataset, there is very little divergence between the train and test dataset except for classes $2$, $7$, and $8$ (corresponding to digits $2$, $7$, and $8$). A similar trend is observed in \textit{FMNIST} dataset with only class $8$ showing high divergence away from train. However, in datasets such as \textit{CIFAR10} and \textit{SVHN} we observe that there is an extreme overlap between the train dataset and test dataset across every class. This could be attributed to the benchmark dataset creation strategy, to an extent. In \textit{SVHN} dataset collection process, all the street view images was collected and annotated together, and then split randomly into train and test datasets. Overall, the existing metrics demonstrate that the test datasets for the existing benchmark datasets are of good quality.

\begin{table*}[htb]
\centering
\footnotesize
\begin{tabular}{|l|l|l|l|l|l|l|l|l|l|l|l|l|l|l|l|}

\toprule
 Dataset & Model & Accuracy & \#samples & 0-100 split & 30-70 split & 50-50 split & 70-30 split  & 100-0 split  \\
\toprule
\multirow{12}{*} {CIFAR10} & \multirow{4}{*} {LeNet} & 70.67 & 100  & 10.90 & 8.60 & 10.40 & 10.00 & 10.00 \\
 &  & 70.67 & 300  & 10.733 & 10.33 & 9.90 & 10.16 & 10.00 \\
 &  & 70.67 & 700  & 9.70 & 9.85 & 10.00 & 9.85 & 10.13 \\
 &  & 70.67 & 1000 & 10.13 & 10.11 & 10.02 & 10.12 & 10.00 \\
\cline{2-9}
& \multirow{4}{*} {ResNet} & 92.40 & 100  & 10.6 & 9.5 & 9.9 & 9.8 & 9.4   \\
&  & 92.40 & 300  & 9.2 & 9.9 & 10.23 & 9.3 & 9.5   \\
 &  & 92.40 & 700  & 10.41 & 9.67 & 9.8 & 9.7 & 9.71  \\
&  & 92.40 & 1000  & 9.71 & 9.87 & 9.73 & 9.91 & 9.70 \\
\cline{2-9}
& \multirow{4}{*} {VGG} & 91.00 & 100  & 9.40 & 11.90 & 13.40 & 12.50 & 13.80  \\ 
 &  & 91.00 & 300  & 9.96 & 11.06 & 12.10 & 13.00 & 13.93   \\
 &  & 91.00 & 700  & 10.20 & 11.28 & 12.24 & 12.37 & 14.30   \\
&  & 91.00 & 1000  & 9.85 & 11.24 & 12.00 & 12.91 & 13.95    \\
\hline
\multirow{12}{*} {CIFAR100} & \multirow{4}{*} {LeNet} & 37.23 & 10  & 1.00 & 1.00 & 5.00 & 4.00 & 4.00   \\ 
&  & 37.23 & 30  & 1.66 & 3.00 & 4.66 & 5.66 & 1.14   \\
&  & 37.23 & 70  & 1.14 & 3.57 & 4.14 & 4.00 & 1.40   \\
&  & 37.23 & 100  & 1.40 & 2.50 & 3.40 & 4.70 & 3.90   \\
\cline{2-9}
 & \multirow{4}{*} {ResNet} & 67.4 & 10  & 0.00 & 0.00 & 1.00 & 0.00 & 0.00   \\ 
 &  & 67.4 & 30  & 0.33 & 0.33 & 1.00 & 0.00 & 0.00   \\
 &  & 67.4 & 70  & 0.71 & 0.28 & 0.28 & 0.00 & 0.00   \\
 &  & 67.4 & 100  & 1.00 & 0.70 & 0.30 & 0.20 & 0.00   \\
\cline{2-9}
 & \multirow{4}{*} {VGG} & 61.38 & 10  & 0.00 & 2.00 & 2.00 & 3.00 & 8.00   \\ 
 &  & 61.38 & 30  & 0.00 & 1.33 & 2.66 & 2.66 & 4.00   \\
&  & 61.38 & 70  & 0.00 & 1.42 & 2.71 & 4.28 & 7.00   \\
 &  & 61.38 & 100 & 0.00 & 2.00 & 3.10 & 4.50 & 5.30   \\
\hline
\multirow{12}{*} {FMNIST} & \multirow{4}{*} {LeNet} & 88.50 & 100  & 8.70 & 32.80 & 48.40 & 63.00 & 86.50   \\
 &  & 88.50 & 300  & 9.60 & 32.46 & 47.80 & 63.86 & 86.66   \\
 &  & 88.50 & 700  & 10.37 & 32.44 & 47.61 & 63.14 & 86.21  \\
 &  & 88.50 & 1000  & 9.88 & 32.67 & 48.16 & 63.21 & 86.11  \\
\cline{2-9}
 & \multirow{4}{*} {ResNet} & 92.58 & 100  & 8.90 & 34.60 & 52.10 & 69.50 & 95.10    \\
 &  & 92.58 & 300  & 10.36 & 35.40 & 52.66 & 69.53 & 94.33    \\
 &  & 92.58 & 700  & 10.14 & 35.44 & 52.52 & 69.08 & 94.61   \\
 &  & 92.58 & 1000  & 10.14 & 35.63 & 52.64 & 69.12 & 94.83   \\
\cline{2-9}
 & \multirow{4}{*} {VGG} & 93.13 & 100  & 10.4 & 24.8 & 37.3 & 46.8 & 62.2    \\
 &  & 93.13 & 300  & 10.06 & 26.03 & 36.76 & 46.5 & 62.7  \\
 &  & 93.13 & 700  & 9.70 & 25.35 & 36.25 & 46.87 & 63.2   \\
 &  & 93.13 & 1000  & 10.14 & 25.82 & 36.45 & 47.6 & 62.3   \\
\hline
\multirow{12}{*} {MNIST} & \multirow{4}{*} {LeNet} & 99.49 & 100  & 9.70 & 36.60 & 54.40 & 72.50 & 99.60   \\
&  & 99.49 & 300  & 9.26 & 35.90 & 54.33 & 72.36 & 99.10    \\
&  & 99.49 & 700  & 10.21 & 36.80 & 54.70 & 72.41 & 99.10    \\
 &  & 99.49 & 1000  & 9.97 & 36.71 & 54.55 & 72.76 & 99.14     \\
\cline{2-9}
 & \multirow{4}{*} {ResNet} & 99.60 & 100  & 9.70 & 35.80 & 54.80 & 73.40 & 100.00   \\
 &  & 99.60 & 300  & 9.96 & 37.06 & 55.50 & 72.56 & 100.00    \\
 &  & 99.60 & 700  & 9.58 & 37.11 & 54.92 & 72.90 & 100.00    \\
 &  & 99.60 & 1000 & 9.99 & 37.20 & 55.21 & 73.13 & 100.00     \\
\cline{2-9}
 & \multirow{4}{*} {VGG} & 99.60 & 100  & 10.4 & 20.3 & 27.4 & 33.6 & 44.4           \\
 &  & 99.60 & 300  & 10.0 & 20.13 & 26.9 & 33.4 & 44.7           \\
&  & 99.60 & 700  & 10.05 & 19.82 & 26.67 & 33.38 & 43.48        \\
 &  & 99.60 & 1000  & 10.06 & 20.03 & 27.14 & 33.21 & 43.56        \\
\hline
\multirow{12}{*} {SVHN} & \multirow{4}{*} {LeNet} & 89.55 & 260  & 10.46 & 14.80 & 18.73 & 22.07 & 27.26       \\
 &  & 89.55 & 781  & 9.88 & 15.53 & 18.59 & 21.69 & 26.82         \\
&  & 89.55 & 1822  & 9.83 & 15.10 & 18.73 & 22.04 & 26.78         \\
 &  & 89.55 & 2603  & 10.21 & 15.01 & 18.28 & 21.84 & 27.16         \\
\cline{2-9}
 & \multirow{4}{*} {ResNet} & 96.48 & 260  & 10.42 & 14.26 & 15.15 & 18.65 & 22.69        \\
 &  & 96.48 & 781  & 10.33 & 13.32 & 15.03 & 18.37 & 21.79         \\
&  & 96.48 & 1822  & 9.79 & 13.21 & 16.02 & 17.99 & 21.78          \\
 &  & 96.48 & 2603  & 10.23 & 13.78 & 16.07 & 18.39 & 22.13          \\
\cline{2-9}
 & \multirow{4}{*} {VGG} & 96.87 & 260  & 9.50 & 21.80 & 29.96 & 37.96 & 52.11         \\
&  & 96.87 & 781  & 10.35 & 22.53 & 30.27 & 38.43 & 51.56         \\
 &  & 96.87 & 1822  & 9.68 & 22.16 & 30.61 & 38.61 & 50.92          \\
 &  & 96.87 & 2603  & 9.84 & 22.18 & 30.85 & 38.82 & 50.46           \\
\bottomrule
\end{tabular}
\caption{Robustness testing of the generated test samples using the proposed systematic approach.}
\label{tab:test_distribution}
\end{table*}

\subsection{RQ2: Quality Analysis using Proposed Metrics}
In this experiment, we study the quality of existing benchmark test sets across the four quality dimensions proposed.  

Equivalence partitioning measures the distribution of test samples across the class labels. Figure 4 illustrates the box-plot representation of equivalence partitioning dimension across the subjects. It is evident from the plot that all test data set except \textit{SVHN}, have sampled test cases equally across all classes is not skewed or biased towards a subset of classes. However \textit{SVHN} test data set is skewed and has over sampled for few classes (namely digits 2, 3, 4 and 5) and under sampled for some classes (namely digits 1, 6, 7, 8, and 9). 

The results shown in Figure \ref{fig:expts}, measures the distribution of the test samples across the three dimensions: border conditioning (BC), centroid partitioning (CP), and pair-wise border conditioning (PBC) for the five datasets across three models. A model should ideally classify the test samples in a similar fashion across class labels. The distribution of test samples classified as centroid or close to centroid should not significantly vary across class labels. In the box-plot where each data point represents the percentage of test samples from each class, we need the percentage to be exactly the same (or) with very less variance. Hence, a model which shows very less variance (smaller the size of the box plot) in the distribution of samples, can be considered again as a robust model, since it is performing equally well across all samples and its learning is not skewed towards certain classes. 

 This property is observed across all models for \textit{CIFAR10}, \textit{MNIST} and \textit{SVHN} test data sets. However, for \textit{CIFAR100} and \textit{FMNIST}, models which are not the top performing models in terms of accuracy exhibit a larger variance. \textit{LeNet} model exhibits variance in all the dimensions for both \textit{FMNIST} and \textit{CIFAR100} data sets and is the lowest performing in terms of accuracy among the three models. Interestingly, \textit{ResNet} is almost a high performing model when compared to \textit{VGG19} for \textit{FMNIST} data set, however the variance in the CP metric is significantly huge. 
 
 From a coverage perspective, very few models exhibit good coverage on the test data set. \textit{LeNet} which in general is performing low from a accuracy perspective across all data sets, is exhibiting a good coverage of test data points. However, highly accurate models \textit{ResNet} and \textit{VGG} for datasets \textit{MNIST} and \textit{SVHN}, are classifying majority of the test data points in the \textit{centroid} region.
 
 Based on this analysis we claim, maybe \textit{VGG19} and \textit{ResNet} models which are high performing models for majority of the data sets, have not been thoroughly tested by sampling enough data points from a coverage perspective.

\subsection{RQ3: Providing better ``Guaranteeable" Performance}
We generate sample test cases based on the algorithm discussed in Section \ref{algo:generation}. For each data set, we generated four different test datasets of samples size 100, 300, 700 and 1000. Further, we sampled test data set across centroid region and boundary in the ratio of 0-100, 30-70, 50-50, 70-30 and 100-0. We generated 20 test data sets for each benchmark data set and model pair. Table \ref{tab:test_distribution} captures accuracy of the models, across these twenty generated test data sets for the five benchmark data sets. 

One observation across all models across all test data sets, is that the accuracy dropped significantly when all the test data samples where in the \textit{boundary} region. The accuracy values were at max 10\% and as low as 0\% (Column (1) in Table \ref{tab:test_distribution}).  As we increase the number of samples from the \textit{centroid} region, the accuracy of the models increased for the generated test data set in \text{FMNIST}, \textit{MNIST} and \textit{SVHN}. 

Interestingly, \textit{LENET} which is the low performing model for dataset \textit{FMNIST} and \textit{MNIST}, has higher accuracy than \textit{VGG19} in all our generated test data sets, indicating that the \textit{LENET} model is more robust than \textit{VGG19}, and the performance is more guaranteed. 

We don't observe the general increase in accuracy numbers for the two data sets \textit{CIFAR10} and \textit{CIFAR100}. The data sets are thumbnail images with very poor quality. Hence the perturbation of features leads to very low inter class variation, resulting in model getting confused across all our test data set.


\begin{figure}[!t]
	\begin{center}
	\includegraphics[width=0.45\textwidth]{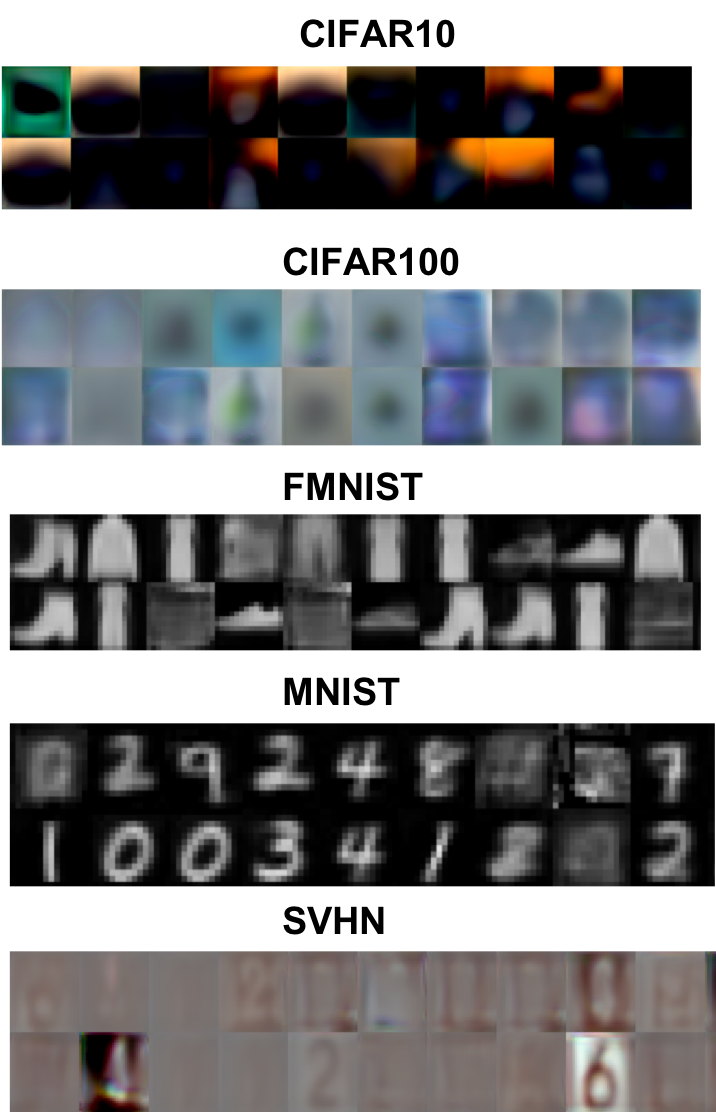}
	\end{center}
	\caption{Sample generated images using the deconvolutional network for different datasets for the VGG-19 model. }
	\label{fig:gen99}
\end{figure}

\subsection{Discussion}


In general what we observe from our experiment is that even though the test data set quality is acceptable from traditional metrics perspective, we showed that from a coverage perspective (which is model specific), the benchmark data sets are not good enough.  Further, models which are low performing in accuracy metric are more robust to our test cases, in comparison to high performing models on the standard bench mark test data set. 

Human consumption of these generated test data points, are  out of scope of this work. Generating a high quality image from feature perturbations is a hard problem and there are some early works with limited success\cite{dosovitskiy2016inverting}. We investigated deconvolutional neural networks to generate the images of the perturbed features and the generated images are shown in Figure~\ref{fig:gen99}. It can be observed that the features perturbed near the centroid region provide a clear visualization in the image space. However, the features perturbed near the boundary generate low quality images that could confuse the classifier. 

The aim of this research to test the functioning of the model and provide a better ``guaranteable" measure by focusing on the coverage of the test dataset. While there are additional branches of DNN model testing for robustness by generating adversarial samples, intentionally attacking or breaking the DNN model is not the primary goal of this research.

\section{Threats to Validity}
In this section we have highlighted few threats to our approach and address the concerns of generalization. 
\begin{enumerate}
    \item \textbf{Generalization}: Our approach is data set modality independent, but specific to \textit{classification} task. We use the features from the last convolution layer of the neural network as input, and this feature extraction as such can be transferred for recurrent neural networks too (for text inputs). However, in our work we have only shown experiment results on image data sets and models trained on these data sets. 
    
    \item \textbf{Layer Selection}: Currently, the test set quality evaluation and the additional test dataset generation is performed for the last convolution layer of the model. We chose this layer, because it gives a good trade off between highly discriminative and generative features.  However, it would be interesting to study the performance of our quality metrics for the feature spaces produced by other layers also similar to debugging layer selection done in \cite{ma2018mode}.
    
    \item \textbf{Validity of generated test cases}: Using the proposed metrics, we sample additional test cases in the feature space and generate the oracle ground truth using metamorphic rules. As these additional points are sampled in the feature space, it is not necessary that all these points have corresponding valid representation in the input image space. Some of the points sampled towards the \textit{centroid} in the feature space, have a clear and legible representation in the input image space while points sampled towards the \textit{boundary} might look visually confusing for a human.
    
    \item \textbf{Threshold values}: The thresholds for choosing the \textit{boundary} and \textit{centroid} regions are heuristically defined through experiments. It may not generalize to all possible CNN models and image datasets and there might be a need to fine-tune these hyperparameters. The obtained results are sensitive to these thresholds and a change in the threshold might potentially result in different results.
    
\end{enumerate} 

\section{Conclusion and Future Work}
In this work, we studied the necessity for testing of DNN models beyond standard accuracy measure. We proposed new metrics to review the quality of test dataset of popular image classification datasets. The metrics were measured on features from penultimate layers of popular models such as \textit{LeNet, ResNet-20,  VGG-19}  on well known datasets like\textit{ MNIST, Fashion-MNIST, CIFAR-10, CIFAR-100}, and \textit{SVHN}. We observe that though \textit{ResNet} performs better than \textit{VGG19} on \textit{CIFAR-10} in terms of accuracy, the coverage of boundary condition testing is comparatively lesser. Further, we generated more test samples guided by the proposed metrics and we observe a drop in accuracy on the previous best models. For \textit{VGG-19}, test dataset consisting only of centroid samples, the accuracy significantly even with $100\%$ samples, thereby validating our approach. 

As part of future work, we consider to improve our algorithm to optimize on the time taken to generate the samples. Our proposed metrics are only for classification tasks, and exploring metrics for test case evaluation for other machine learning tasks like segmentation, regression and modalities like text can be an interesting future work. 


%
\bibliographystyle{ACM-Reference-Format}
\bibliography{sample-base}


\begin{thebibliography}{33}


\ifx \showCODEN    \undefined \def \showCODEN     #1{\unskip}     \fi
\ifx \showDOI      \undefined \def \showDOI       #1{#1}\fi
\ifx \showISBNx    \undefined \def \showISBNx     #1{\unskip}     \fi
\ifx \showISBNxiii \undefined \def \showISBNxiii  #1{\unskip}     \fi
\ifx \showISSN     \undefined \def \showISSN      #1{\unskip}     \fi
\ifx \showLCCN     \undefined \def \showLCCN      #1{\unskip}     \fi
\ifx \shownote     \undefined \def \shownote      #1{#1}          \fi
\ifx \showarticletitle \undefined \def \showarticletitle #1{#1}   \fi
\ifx \showURL      \undefined \def \showURL       {\relax}        \fi
\providecommand\bibfield[2]{#2}
\providecommand\bibinfo[2]{#2}
\providecommand\natexlab[1]{#1}
\providecommand\showeprint[2][]{arXiv:#2}

\bibitem[\protect\citeauthoryear{Angelova, Krizhevsky, Vanhoucke, Ogale, and
  Ferguson}{Angelova et~al\mbox{.}}{2015}]%
        {angelova2015real}
\bibfield{author}{\bibinfo{person}{Anelia Angelova}, \bibinfo{person}{Alex
  Krizhevsky}, \bibinfo{person}{Vincent Vanhoucke}, \bibinfo{person}{Abhijit~S
  Ogale}, {and} \bibinfo{person}{Dave Ferguson}.}
  \bibinfo{year}{2015}\natexlab{}.
\newblock \showarticletitle{Real-Time Pedestrian Detection with Deep Network
  Cascades.}. In \bibinfo{booktitle}{\emph{BMVC}}, Vol.~\bibinfo{volume}{2}.
  \bibinfo{pages}{4}.
\newblock


\bibitem[\protect\citeauthoryear{Bojarski, Del~Testa, Dworakowski, Firner,
  Flepp, Goyal, Jackel, Monfort, Muller, Zhang, et~al\mbox{.}}{Bojarski
  et~al\mbox{.}}{2016}]%
        {bojarski2016end}
\bibfield{author}{\bibinfo{person}{Mariusz Bojarski}, \bibinfo{person}{Davide
  Del~Testa}, \bibinfo{person}{Daniel Dworakowski}, \bibinfo{person}{Bernhard
  Firner}, \bibinfo{person}{Beat Flepp}, \bibinfo{person}{Prasoon Goyal},
  \bibinfo{person}{Lawrence~D Jackel}, \bibinfo{person}{Mathew Monfort},
  \bibinfo{person}{Urs Muller}, \bibinfo{person}{Jiakai Zhang},
  {et~al\mbox{.}}} \bibinfo{year}{2016}\natexlab{}.
\newblock \showarticletitle{End to end learning for self-driving cars}.
\newblock \bibinfo{journal}{\emph{arXiv preprint arXiv:1604.07316}}
  (\bibinfo{year}{2016}).
\newblock


\bibitem[\protect\citeauthoryear{Carlini and Wagner}{Carlini and
  Wagner}{2017}]%
        {carlini2017towards}
\bibfield{author}{\bibinfo{person}{Nicholas Carlini} {and}
  \bibinfo{person}{David Wagner}.} \bibinfo{year}{2017}\natexlab{}.
\newblock \showarticletitle{Towards evaluating the robustness of neural
  networks}. In \bibinfo{booktitle}{\emph{Security and Privacy (SP), 2017 IEEE
  Symposium on}}. IEEE, \bibinfo{pages}{39--57}.
\newblock


\bibitem[\protect\citeauthoryear{Dosovitskiy and Brox}{Dosovitskiy and
  Brox}{2016}]%
        {dosovitskiy2016inverting}
\bibfield{author}{\bibinfo{person}{Alexey Dosovitskiy} {and}
  \bibinfo{person}{Thomas Brox}.} \bibinfo{year}{2016}\natexlab{}.
\newblock \showarticletitle{Inverting visual representations with convolutional
  networks}. In \bibinfo{booktitle}{\emph{Proceedings of the IEEE Conference on
  Computer Vision and Pattern Recognition}}. \bibinfo{pages}{4829--4837}.
\newblock


\bibitem[\protect\citeauthoryear{Esteva, Kuprel, Novoa, Ko, Swetter, Blau, and
  Thrun}{Esteva et~al\mbox{.}}{2017}]%
        {esteva2017dermatologist}
\bibfield{author}{\bibinfo{person}{Andre Esteva}, \bibinfo{person}{Brett
  Kuprel}, \bibinfo{person}{Roberto~A Novoa}, \bibinfo{person}{Justin Ko},
  \bibinfo{person}{Susan~M Swetter}, \bibinfo{person}{Helen~M Blau}, {and}
  \bibinfo{person}{Sebastian Thrun}.} \bibinfo{year}{2017}\natexlab{}.
\newblock \showarticletitle{Dermatologist-level classification of skin cancer
  with deep neural networks}.
\newblock \bibinfo{journal}{\emph{Nature}} \bibinfo{volume}{542},
  \bibinfo{number}{7639} (\bibinfo{year}{2017}), \bibinfo{pages}{115}.
\newblock


\bibitem[\protect\citeauthoryear{Fei-Fei}{Fei-Fei}{2010}]%
        {fei2010imagenet}
\bibfield{author}{\bibinfo{person}{Li Fei-Fei}.}
  \bibinfo{year}{2010}\natexlab{}.
\newblock \showarticletitle{ImageNet: crowdsourcing, benchmarking \& other cool
  things}. In \bibinfo{booktitle}{\emph{CMU VASC Seminar}},
  Vol.~\bibinfo{volume}{16}. \bibinfo{pages}{18--25}.
\newblock


\bibitem[\protect\citeauthoryear{Hayhurst, Veerhusen, Chilenski, and
  Rierson}{Hayhurst et~al\mbox{.}}{2001}]%
        {hayhurst2001practical}
\bibfield{author}{\bibinfo{person}{Kelly~J Hayhurst}, \bibinfo{person}{Dan~S
  Veerhusen}, \bibinfo{person}{John~J Chilenski}, {and}
  \bibinfo{person}{Leanna~K Rierson}.} \bibinfo{year}{2001}\natexlab{}.
\newblock \showarticletitle{A practical tutorial on modified condition/decision
  coverage}.
\newblock  (\bibinfo{year}{2001}).
\newblock


\bibitem[\protect\citeauthoryear{He, Zhang, Ren, and Sun}{He
  et~al\mbox{.}}{2016}]%
        {he2016deep}
\bibfield{author}{\bibinfo{person}{Kaiming He}, \bibinfo{person}{Xiangyu
  Zhang}, \bibinfo{person}{Shaoqing Ren}, {and} \bibinfo{person}{Jian Sun}.}
  \bibinfo{year}{2016}\natexlab{}.
\newblock \showarticletitle{Deep residual learning for image recognition}. In
  \bibinfo{booktitle}{\emph{Proceedings of the IEEE conference on computer
  vision and pattern recognition}}. \bibinfo{pages}{770--778}.
\newblock


\bibitem[\protect\citeauthoryear{Huval, Wang, Tandon, Kiske, Song,
  Pazhayampallil, Andriluka, Rajpurkar, Migimatsu, Cheng-Yue,
  et~al\mbox{.}}{Huval et~al\mbox{.}}{2015}]%
        {huval2015empirical}
\bibfield{author}{\bibinfo{person}{Brody Huval}, \bibinfo{person}{Tao Wang},
  \bibinfo{person}{Sameep Tandon}, \bibinfo{person}{Jeff Kiske},
  \bibinfo{person}{Will Song}, \bibinfo{person}{Joel Pazhayampallil},
  \bibinfo{person}{Mykhaylo Andriluka}, \bibinfo{person}{Pranav Rajpurkar},
  \bibinfo{person}{Toki Migimatsu}, \bibinfo{person}{Royce Cheng-Yue},
  {et~al\mbox{.}}} \bibinfo{year}{2015}\natexlab{}.
\newblock \showarticletitle{An empirical evaluation of deep learning on highway
  driving}.
\newblock \bibinfo{journal}{\emph{arXiv preprint arXiv:1504.01716}}
  (\bibinfo{year}{2015}).
\newblock


\bibitem[\protect\citeauthoryear{Krizhevsky and Hinton}{Krizhevsky and
  Hinton}{2009}]%
        {krizhevsky2009learning}
\bibfield{author}{\bibinfo{person}{Alex Krizhevsky} {and}
  \bibinfo{person}{Geoffrey Hinton}.} \bibinfo{year}{2009}\natexlab{}.
\newblock \bibinfo{booktitle}{\emph{Learning multiple layers of features from
  tiny images}}.
\newblock \bibinfo{type}{{T}echnical {R}eport}.
  \bibinfo{institution}{Citeseer}.
\newblock


\bibitem[\protect\citeauthoryear{LeCun et~al\mbox{.}}{LeCun
  et~al\mbox{.}}{2015}]%
        {lecun2015lenet}
\bibfield{author}{\bibinfo{person}{Yann LeCun} {et~al\mbox{.}}}
  \bibinfo{year}{2015}\natexlab{}.
\newblock \showarticletitle{LeNet-5, convolutional neural networks}.
\newblock \bibinfo{journal}{\emph{URL: http://yann. lecun. com/exdb/lenet}}
  \bibinfo{volume}{20} (\bibinfo{year}{2015}).
\newblock


\bibitem[\protect\citeauthoryear{LeCun, Bottou, Bengio, Haffner,
  et~al\mbox{.}}{LeCun et~al\mbox{.}}{1998}]%
        {lecun1998gradient}
\bibfield{author}{\bibinfo{person}{Yann LeCun}, \bibinfo{person}{L{\'e}on
  Bottou}, \bibinfo{person}{Yoshua Bengio}, \bibinfo{person}{Patrick Haffner},
  {et~al\mbox{.}}} \bibinfo{year}{1998}\natexlab{}.
\newblock \showarticletitle{Gradient-based learning applied to document
  recognition}.
\newblock \bibinfo{journal}{\emph{Proc. IEEE}} \bibinfo{volume}{86},
  \bibinfo{number}{11} (\bibinfo{year}{1998}), \bibinfo{pages}{2278--2324}.
\newblock


\bibitem[\protect\citeauthoryear{Ma, Juefei-Xu, Sun, Chen, Su, Zhang, Xue, Li,
  Li, Liu, et~al\mbox{.}}{Ma et~al\mbox{.}}{2018a}]%
        {ma2018deepgauge}
\bibfield{author}{\bibinfo{person}{Lei Ma}, \bibinfo{person}{Felix Juefei-Xu},
  \bibinfo{person}{Jiyuan Sun}, \bibinfo{person}{Chunyang Chen},
  \bibinfo{person}{Ting Su}, \bibinfo{person}{Fuyuan Zhang},
  \bibinfo{person}{Minhui Xue}, \bibinfo{person}{Bo Li}, \bibinfo{person}{Li
  Li}, \bibinfo{person}{Yang Liu}, {et~al\mbox{.}}}
  \bibinfo{year}{2018}\natexlab{a}.
\newblock \showarticletitle{DeepGauge: Comprehensive and Multi-Granularity
  Testing Criteria for Gauging the Robustness of Deep Learning Systems}.
\newblock \bibinfo{journal}{\emph{arXiv preprint arXiv:1803.07519}}
  (\bibinfo{year}{2018}).
\newblock


\bibitem[\protect\citeauthoryear{Ma, Zhang, Sun, Xue, Li, Juefei-Xu, Xie, Li,
  Liu, Zhao, et~al\mbox{.}}{Ma et~al\mbox{.}}{2018c}]%
        {ma2018deepmutation}
\bibfield{author}{\bibinfo{person}{Lei Ma}, \bibinfo{person}{Fuyuan Zhang},
  \bibinfo{person}{Jiyuan Sun}, \bibinfo{person}{Minhui Xue},
  \bibinfo{person}{Bo Li}, \bibinfo{person}{Felix Juefei-Xu},
  \bibinfo{person}{Chao Xie}, \bibinfo{person}{Li Li}, \bibinfo{person}{Yang
  Liu}, \bibinfo{person}{Jianjun Zhao}, {et~al\mbox{.}}}
  \bibinfo{year}{2018}\natexlab{c}.
\newblock \showarticletitle{DeepMutation: Mutation Testing of Deep Learning
  Systems}.
\newblock \bibinfo{journal}{\emph{arXiv preprint arXiv:1805.05206}}
  (\bibinfo{year}{2018}).
\newblock


\bibitem[\protect\citeauthoryear{Ma, Liu, Lee, Zhang, and Grama}{Ma
  et~al\mbox{.}}{2018b}]%
        {ma2018mode}
\bibfield{author}{\bibinfo{person}{Shiqing Ma}, \bibinfo{person}{Yingqi Liu},
  \bibinfo{person}{Wen-Chuan Lee}, \bibinfo{person}{Xiangyu Zhang}, {and}
  \bibinfo{person}{Ananth Grama}.} \bibinfo{year}{2018}\natexlab{b}.
\newblock \showarticletitle{MODE: automated neural network model debugging via
  state differential analysis and input selection}. In
  \bibinfo{booktitle}{\emph{Proceedings of the 2018 26th ACM Joint Meeting on
  European Software Engineering Conference and Symposium on the Foundations of
  Software Engineering}}. ACM, \bibinfo{pages}{175--186}.
\newblock


\bibitem[\protect\citeauthoryear{Medford}{Medford}{2016}]%
        {GoogleFinanceMonitise20140426}
\bibfield{author}{\bibinfo{person}{Ron Medford}.}
  \bibinfo{year}{2016}\natexlab{}.
\newblock \bibinfo{title}{Google Auto Waymo Disengagement Report for Autonomous
  Driving}.
\newblock
  \bibinfo{howpublished}{\url{www.dmv.ca.gov/portal/wcm/connect/946b3502-c959-4e3b-b119-91319c27788f/GoogleAutoWaymo_disengage_report_2016.pdf?MOD=AJPERES}}.
\newblock


\bibitem[\protect\citeauthoryear{Milletari, Navab, and Ahmadi}{Milletari
  et~al\mbox{.}}{2016}]%
        {milletari2016v}
\bibfield{author}{\bibinfo{person}{Fausto Milletari}, \bibinfo{person}{Nassir
  Navab}, {and} \bibinfo{person}{Seyed-Ahmad Ahmadi}.}
  \bibinfo{year}{2016}\natexlab{}.
\newblock \showarticletitle{V-net: Fully convolutional neural networks for
  volumetric medical image segmentation}. In \bibinfo{booktitle}{\emph{3D
  Vision (3DV), 2016 Fourth International Conference on}}. IEEE,
  \bibinfo{pages}{565--571}.
\newblock


\bibitem[\protect\citeauthoryear{Moosavi-Dezfooli, Fawzi, Fawzi, and
  Frossard}{Moosavi-Dezfooli et~al\mbox{.}}{2017}]%
        {moosavi2017universal}
\bibfield{author}{\bibinfo{person}{Seyed-Mohsen Moosavi-Dezfooli},
  \bibinfo{person}{Alhussein Fawzi}, \bibinfo{person}{Omar Fawzi}, {and}
  \bibinfo{person}{Pascal Frossard}.} \bibinfo{year}{2017}\natexlab{}.
\newblock \showarticletitle{Universal adversarial perturbations}.
\newblock \bibinfo{journal}{\emph{arXiv preprint}} (\bibinfo{year}{2017}).
\newblock


\bibitem[\protect\citeauthoryear{Netzer, Wang, Coates, Bissacco, Wu, and
  Ng}{Netzer et~al\mbox{.}}{2011}]%
        {netzer2011reading}
\bibfield{author}{\bibinfo{person}{Yuval Netzer}, \bibinfo{person}{Tao Wang},
  \bibinfo{person}{Adam Coates}, \bibinfo{person}{Alessandro Bissacco},
  \bibinfo{person}{Bo Wu}, {and} \bibinfo{person}{Andrew~Y Ng}.}
  \bibinfo{year}{2011}\natexlab{}.
\newblock \showarticletitle{Reading digits in natural images with unsupervised
  feature learning}.
\newblock  (\bibinfo{year}{2011}).
\newblock


\bibitem[\protect\citeauthoryear{Nguyen, Yosinski, and Clune}{Nguyen
  et~al\mbox{.}}{2015}]%
        {nguyen2015deep}
\bibfield{author}{\bibinfo{person}{Anh Nguyen}, \bibinfo{person}{Jason
  Yosinski}, {and} \bibinfo{person}{Jeff Clune}.}
  \bibinfo{year}{2015}\natexlab{}.
\newblock \showarticletitle{Deep neural networks are easily fooled: High
  confidence predictions for unrecognizable images}. In
  \bibinfo{booktitle}{\emph{Proceedings of the IEEE Conference on Computer
  Vision and Pattern Recognition}}. \bibinfo{pages}{427--436}.
\newblock


\bibitem[\protect\citeauthoryear{Odena and Goodfellow}{Odena and
  Goodfellow}{2018}]%
        {odena2018tensorfuzz}
\bibfield{author}{\bibinfo{person}{Augustus Odena} {and} \bibinfo{person}{Ian
  Goodfellow}.} \bibinfo{year}{2018}\natexlab{}.
\newblock \showarticletitle{TensorFuzz: Debugging Neural Networks with
  Coverage-Guided Fuzzing}.
\newblock \bibinfo{journal}{\emph{arXiv preprint arXiv:1807.10875}}
  (\bibinfo{year}{2018}).
\newblock


\bibitem[\protect\citeauthoryear{Pei, Cao, Yang, and Jana}{Pei
  et~al\mbox{.}}{2017}]%
        {pei2017deepxplore}
\bibfield{author}{\bibinfo{person}{Kexin Pei}, \bibinfo{person}{Yinzhi Cao},
  \bibinfo{person}{Junfeng Yang}, {and} \bibinfo{person}{Suman Jana}.}
  \bibinfo{year}{2017}\natexlab{}.
\newblock \showarticletitle{Deepxplore: Automated whitebox testing of deep
  learning systems}. In \bibinfo{booktitle}{\emph{Proceedings of the 26th
  Symposium on Operating Systems Principles}}. ACM, \bibinfo{pages}{1--18}.
\newblock


\bibitem[\protect\citeauthoryear{Shin, Roth, Gao, Lu, Xu, Nogues, Yao, Mollura,
  and Summers}{Shin et~al\mbox{.}}{2016}]%
        {shin2016deep}
\bibfield{author}{\bibinfo{person}{Hoo-Chang Shin}, \bibinfo{person}{Holger~R
  Roth}, \bibinfo{person}{Mingchen Gao}, \bibinfo{person}{Le Lu},
  \bibinfo{person}{Ziyue Xu}, \bibinfo{person}{Isabella Nogues},
  \bibinfo{person}{Jianhua Yao}, \bibinfo{person}{Daniel Mollura}, {and}
  \bibinfo{person}{Ronald~M Summers}.} \bibinfo{year}{2016}\natexlab{}.
\newblock \showarticletitle{Deep convolutional neural networks for
  computer-aided detection: CNN architectures, dataset characteristics and
  transfer learning}.
\newblock \bibinfo{journal}{\emph{IEEE transactions on medical imaging}}
  \bibinfo{volume}{35}, \bibinfo{number}{5} (\bibinfo{year}{2016}),
  \bibinfo{pages}{1285--1298}.
\newblock


\bibitem[\protect\citeauthoryear{Simonyan and Zisserman}{Simonyan and
  Zisserman}{2014}]%
        {simonyan2014very}
\bibfield{author}{\bibinfo{person}{Karen Simonyan} {and}
  \bibinfo{person}{Andrew Zisserman}.} \bibinfo{year}{2014}\natexlab{}.
\newblock \showarticletitle{Very deep convolutional networks for large-scale
  image recognition}.
\newblock \bibinfo{journal}{\emph{arXiv preprint arXiv:1409.1556}}
  (\bibinfo{year}{2014}).
\newblock


\bibitem[\protect\citeauthoryear{Storkey}{Storkey}{2009}]%
        {storkey2009training}
\bibfield{author}{\bibinfo{person}{Amos Storkey}.}
  \bibinfo{year}{2009}\natexlab{}.
\newblock \showarticletitle{When training and test sets are different:
  characterizing learning transfer}.
\newblock \bibinfo{journal}{\emph{Dataset shift in machine learning}}
  (\bibinfo{year}{2009}), \bibinfo{pages}{3--28}.
\newblock


\bibitem[\protect\citeauthoryear{Sun, Huang, and Kroening}{Sun
  et~al\mbox{.}}{2018a}]%
        {sun2018testing}
\bibfield{author}{\bibinfo{person}{Youcheng Sun}, \bibinfo{person}{Xiaowei
  Huang}, {and} \bibinfo{person}{Daniel Kroening}.}
  \bibinfo{year}{2018}\natexlab{a}.
\newblock \showarticletitle{Testing Deep Neural Networks}.
\newblock \bibinfo{journal}{\emph{arXiv preprint arXiv:1803.04792}}
  (\bibinfo{year}{2018}).
\newblock


\bibitem[\protect\citeauthoryear{Sun, Wu, Ruan, Huang, Kwiatkowska, and
  Kroening}{Sun et~al\mbox{.}}{2018b}]%
        {sun2018concolic}
\bibfield{author}{\bibinfo{person}{Youcheng Sun}, \bibinfo{person}{Min Wu},
  \bibinfo{person}{Wenjie Ruan}, \bibinfo{person}{Xiaowei Huang},
  \bibinfo{person}{Marta Kwiatkowska}, {and} \bibinfo{person}{Daniel
  Kroening}.} \bibinfo{year}{2018}\natexlab{b}.
\newblock \showarticletitle{Concolic Testing for Deep Neural Networks}.
\newblock \bibinfo{journal}{\emph{arXiv preprint arXiv:1805.00089}}
  (\bibinfo{year}{2018}).
\newblock


\bibitem[\protect\citeauthoryear{Szegedy, Zaremba, Sutskever, Bruna, Erhan,
  Goodfellow, and Fergus}{Szegedy et~al\mbox{.}}{2013}]%
        {szegedy2013intriguing}
\bibfield{author}{\bibinfo{person}{Christian Szegedy},
  \bibinfo{person}{Wojciech Zaremba}, \bibinfo{person}{Ilya Sutskever},
  \bibinfo{person}{Joan Bruna}, \bibinfo{person}{Dumitru Erhan},
  \bibinfo{person}{Ian Goodfellow}, {and} \bibinfo{person}{Rob Fergus}.}
  \bibinfo{year}{2013}\natexlab{}.
\newblock \showarticletitle{Intriguing properties of neural networks}.
\newblock \bibinfo{journal}{\emph{arXiv preprint arXiv:1312.6199}}
  (\bibinfo{year}{2013}).
\newblock


\bibitem[\protect\citeauthoryear{Tian, Pei, Jana, and Ray}{Tian
  et~al\mbox{.}}{2018}]%
        {tian2018deeptest}
\bibfield{author}{\bibinfo{person}{Yuchi Tian}, \bibinfo{person}{Kexin Pei},
  \bibinfo{person}{Suman Jana}, {and} \bibinfo{person}{Baishakhi Ray}.}
  \bibinfo{year}{2018}\natexlab{}.
\newblock \showarticletitle{Deeptest: Automated testing of
  deep-neural-network-driven autonomous cars}. In
  \bibinfo{booktitle}{\emph{Proceedings of the 40th International Conference on
  Software Engineering}}. ACM, \bibinfo{pages}{303--314}.
\newblock


\bibitem[\protect\citeauthoryear{Torralba and Efros}{Torralba and
  Efros}{2011}]%
        {torralba2011unbiased}
\bibfield{author}{\bibinfo{person}{Antonio Torralba} {and}
  \bibinfo{person}{Alexei~A Efros}.} \bibinfo{year}{2011}\natexlab{}.
\newblock \showarticletitle{Unbiased look at dataset bias}. In
  \bibinfo{booktitle}{\emph{Computer Vision and Pattern Recognition (CVPR),
  2011 IEEE Conference on}}. IEEE, \bibinfo{pages}{1521--1528}.
\newblock


\bibitem[\protect\citeauthoryear{Turhan}{Turhan}{2012}]%
        {turhan2012dataset}
\bibfield{author}{\bibinfo{person}{Burak Turhan}.}
  \bibinfo{year}{2012}\natexlab{}.
\newblock \showarticletitle{On the dataset shift problem in software
  engineering prediction models}.
\newblock \bibinfo{journal}{\emph{Empirical Software Engineering}}
  \bibinfo{volume}{17}, \bibinfo{number}{1-2} (\bibinfo{year}{2012}),
  \bibinfo{pages}{62--74}.
\newblock


\bibitem[\protect\citeauthoryear{Xiao, Rasul, and Vollgraf}{Xiao
  et~al\mbox{.}}{2017}]%
        {xiao2017fashion}
\bibfield{author}{\bibinfo{person}{Han Xiao}, \bibinfo{person}{Kashif Rasul},
  {and} \bibinfo{person}{Roland Vollgraf}.} \bibinfo{year}{2017}\natexlab{}.
\newblock \showarticletitle{Fashion-mnist: a novel image dataset for
  benchmarking machine learning algorithms}.
\newblock \bibinfo{journal}{\emph{arXiv preprint arXiv:1708.07747}}
  (\bibinfo{year}{2017}).
\newblock


\bibitem[\protect\citeauthoryear{Zhang, Kahn, Levine, and Abbeel}{Zhang
  et~al\mbox{.}}{2016}]%
        {zhang2016learning}
\bibfield{author}{\bibinfo{person}{Tianhao Zhang}, \bibinfo{person}{Gregory
  Kahn}, \bibinfo{person}{Sergey Levine}, {and} \bibinfo{person}{Pieter
  Abbeel}.} \bibinfo{year}{2016}\natexlab{}.
\newblock \showarticletitle{Learning deep control policies for autonomous
  aerial vehicles with mpc-guided policy search}. In
  \bibinfo{booktitle}{\emph{Robotics and Automation (ICRA), 2016 IEEE
  International Conference on}}. IEEE, \bibinfo{pages}{528--535}.
\newblock


\end{thebibliography}

%
\appendix

\end{document}